\documentclass[sigconf]{acmart}
\usepackage{duckuments}     %
\usepackage{amsmath,graphicx,amsthm}
\usepackage[ruled,linesnumbered]{algorithm2e}
\usepackage{algorithmic}
\usepackage{listings}
\usepackage{cuted}
\usepackage{paralist} %
\usepackage{xfrac}       %
\usepackage{color}
\usepackage{rotating}
\usepackage{float}
\usepackage{hyperref}
\usepackage[mathscr]{eucal} %
\usepackage{amsbsy} %
\usepackage{subfigure}
\usepackage{subcaption}
\usepackage{booktabs}       %
\usepackage{tcolorbox}
\usepackage{multirow}
\usepackage{todonotes}
\usepackage{epstopdf}
\usepackage{balance}
\usepackage{tablefootnote}
\usepackage{footnote}
\usepackage[symbol]{footmisc}
\usepackage{printlen}
\usepackage{empheq} %
\usepackage{moresize}
\usepackage{enumitem}
\usepackage{graphicx}
\usepackage[utf8]{inputenc} %
\usepackage{nicematrix}
\usepackage{wrapfig} %
\usepackage[capitalize,noabbrev,nameinlink]{cleveref}
\setlist{nolistsep}
\usepackage{pifont} %
\usepackage{xfp}
\usepackage{anyfontsize}
\usepackage{layouts}
\usepackage{microtype}      %

\begin{document}

\newcommand{\method}{\textsc{GpTen}}{}
\newcommand{\dataset}{{\bf G}PT {\bf R}edd{\bf i}t {\bf D}ataset (GRiD)}{}

\newcommand{\datasetlink}{\url{https://github.com/madlab-ucr/GriD}}

\copyrightyear{2024} 
\acmYear{2024} 
\setcopyright{rightsretained} 
\acmConference[WWW '24 Companion]{Companion Proceedings of the ACM Web Conference 2024}{May 13--17, 2024}{Singapore, Singapore}
\acmBooktitle{Companion Proceedings of the ACM Web Conference 2024 (WWW '24 Companion), May 13--17, 2024, Singapore, Singapore}\acmDOI{10.1145/3589335.3651513}
\acmISBN{979-8-4007-0172-6/24/05}

\title{GPT-generated Text Detection: Benchmark Dataset and Tensor-based Detection Method}

\author{Zubair Qazi}%
\email{zqazi004@ucr.edu}%
\affiliation{%
  \institution{University of California, Riverside}
  \city{Riverside}
  \state{CA}
  \country{USA}
}

\author{William Shiao}
\orcid{0000-0001-5813-2266}
\email{wshia002@ucr.edu}
\affiliation{%
  \institution{University of California, Riverside}
  \city{Riverside}
  \state{CA}
  \country{USA}
}

\author{Evangelos E. Papalexakis}
\email{epapalex@cs.ucr.edu}
\orcid{0000-0002-3411-8483}
\affiliation{%
  \institution{University of California, Riverside}
  \city{Riverside}
  \state{CA}
  \country{USA}
}

\begin{abstract}
As natural language models like ChatGPT become increasingly prevalent in applications and services, the need for robust and accurate methods to detect their output is of paramount importance. In this paper, we present  \dataset, a novel Generative Pretrained Transformer (GPT)-generated text detection dataset designed to assess the performance of detection models in identifying generated responses from ChatGPT. The dataset consists of a diverse collection of context-prompt pairs based on Reddit, with human-generated and ChatGPT-generated responses. We provide an analysis of the dataset's characteristics, including linguistic diversity, context complexity, and response quality. To showcase the dataset's utility, we benchmark several detection methods on it, demonstrating their efficacy in distinguishing between human and ChatGPT-generated responses. This dataset serves as a resource for evaluating and advancing detection techniques in the context of ChatGPT and contributes to the ongoing efforts to ensure responsible and trustworthy AI-driven communication on the internet. Finally, we propose \method{}, a novel tensor-based GPT text detection method that is semi-supervised in nature since it only has access to human-generated text and performs on par with fully-supervised baselines.
\end{abstract}
\maketitle              %
\section{Introduction}

Detection of Generative Pretrained Transformer (GPT)-generated content has gained significant relevance with the proliferation of large language models over the internet. These models, including GPT-3, produce human-like text that can be seamlessly integrated into various applications and platforms \cite{gpt3paper}. However, the potential misuse of such generated content for misinformation, spam, or other malicious purposes has raised the importance of detecting GPT-generated text \cite{crothers2023machine}. In diverse contexts like social media, customer service, and content generation, distinguishing between human-authored and AI-generated text has become crucial to maintaining trust, security, and the integrity of online discourse. Detecting GPT-generated text helps to mitigate the risk of spreading disinformation, ensures ethical AI use, and enhances content quality and reliability in applications harnessing AI language models.

There are quite a few approaches that exist for GPT detection \cite{crothers2023machine}. We can sort a majority of the approaches into a few categories: traditional supervised machine learning, deep learning methods, transfer learning methods, and unsupervised methods.

Traditional supervised machine learning methods have been extensively employed for GPT detection \cite{traditionalml}. These approaches leverage labeled datasets, where human-generated and machine-generated text samples are used to train classifiers. One of the key advantages of this approach is its interpretability, as it allows for the examination of features used by classifiers to make predictions. However, traditional supervised methods often require substantial manual annotation efforts to create labeled datasets, which can be time-consuming and resource-intensive. Additionally, they require large amounts of training data, with the risk of overfitting, and may struggle to adapt to evolving GPT models and the diverse ways in which they are employed, making them less effective in dynamic environments such as the modern web.

Deep learning methods, on the other hand, are prominent for their ability to automatically learn complex patterns from data \cite{traditionalml}. These methods, such as neural networks, can effectively capture the nuanced characteristics of GPT-generated text. Deep learning models excel in handling unstructured data, but they tend to be data-hungry and may demand large training datasets for optimal performance. They are widely used due to their robustness and adaptability, especially when substantial labeled data is available.

Transfer learning methods have emerged as a highly practical solution for GPT detection. By leveraging pre-trained models and fine-tuning them on specific tasks, transfer learning allows for efficient use of available resources while inheriting the knowledge and capabilities of the pre-trained models, which can be particularly advantageous in scenarios with limited training data \cite{ruder2019transfer}. However, transfer learning methods may not always generalize well to diverse GPT variants and applications, which can restrict their usefulness.

Unsupervised methods represent a different paradigm, where GPT detection is achieved without the need for labeled data. These methods rely on various statistical and linguistic cues to identify machine-generated content \cite{haj2019towards}. Unsupervised approaches are advantageous for their independence from labeled datasets but can be less accurate and robust compared to supervised or deep learning methods. They are less commonly used in practice due to their limitations, especially in the face of evolving GPT models and sophisticated adversarial techniques.

Traditional supervised machine learning and deep learning methods are commonly favored for their accuracy and adaptability, while transfer learning methods offer a pragmatic balance between data efficiency and effectiveness. Unsupervised methods, although less commonly used, offer a label-free alternative but may lag in terms of accuracy and robustness, especially in complex and evolving GPT environments.
Our contributions in this paper are:
\begin{itemize}
    \item {\bf Dataset}: we present \dataset, a dataset designed and built for GPT detection. We make our dataset publicly available.%
    \item {\bf Novel Method}: we propose \method{}: a novel {\em semi-supervised} tensor-based method with comparable results to existing {\em fully supervised} approaches for GPT detection
    \item {\bf Experimental Evaluation}: we extensively evaluate how state-of-the-art existing approaches behave on our dataset.
\end{itemize}
Our dataset and implementation are publicly available at \datasetlink.

\section{GPT Reddit Dataset Description}

The \dataset~ is a comprehensive collection of text data obtained from two distinct sources: Reddit and the OpenAI API. It is structured to encompass a total of 6513 samples, further categorized into two primary groups: 1368 samples represent text content generated by the GPT-3.5-turbo model, whereas 5145 samples denote text authored by human contributors. Each individual sample within the dataset is labeled to indicate its source of generation, differentiating between GPT-generated and human-generated text. In order to minimize the potential mixture of GPT-generated data with human-generated data, all data from human contributors is dated from October 31 2022 or earlier, which is the official release date of the ChatGPT web application.

The dataset is stored in a structured CSV (Comma-Separated Values) format. Each line in the CSV file consists of a data sample and its corresponding label. The GPT-generated data contained within this dataset is a result of interactions with the GPT-3.5-turbo model provided by the OpenAI API. To solicit responses from the model, a specific prompt was employed: ``You are a frequent user of the subreddits <subreddit\_names>. Answer anything relevant.'' The model's responses to these prompts are integral to the GPT-generated portion of the dataset.
In order to promote further research in this direction, we make this dataset publicly accessible to the research community on GitHub\footnote{\datasetlink}.

\subsection{Dataset Collection}
The human-generated data is gathered from Reddit using the PRAW Python library and GPT-generated content is gathered from the OpenAI API. We sourced the Reddit data from three different subreddits: AskHistorians, AskScience, and ExplainLikeImFive. In order to consider a post from each subreddit, they had to satisfy all of the following criteria:

\begin{enumerate}
    \item The post must be dated before November 2022.
    \item The post must have at least a score (upvotes) of 1000.
    \item The post does not contain adult content. 
    \item The post is in English. 
    \item The post title is formatted as a question.
    \item The post itself cannot be deleted.
\end{enumerate}

\noindent We can justify each criteria separately. The following is a justification for each criteria: 
\begin{enumerate}
    \item ChatGPT was officially released to the public in November 2022. In order to ensure minimal representation of GPT in the human-generated data, we only considered posts before that date. 
    \item We only consider the top posts from each subreddit. Posts with at least 1000 upvotes were generally appropriate for the dataset.
    \item Since the dataset is for academic research purposes, we avoid any adult posts. 
    \item The dataset is only intended to be comprised of English based content at this time. 
    \item Each post title is fed into GPT as a prompt, so in order to reduce noise and increase consistency between the human-generated and GPT generated content.
    \item Some posts on Reddit are deleted by moderators of the subreddit, but their metadata still exists on the subreddit. For fairness, we filter out these posts from the dataset.
\end{enumerate}

\noindent For the current dataset, we gather up to the top 500 posts from each subreddit which satisfy the above criteria. For each post, we gather up to 5 of the top comments based on score (upvotes), and then feed the post title into GPT and store the corresponding response. Each comment only needs to satisfy a simpler criteria to be considered: 
(1) The comment is in English; (2) The comment is not deleted.

\subsection{Dataset Processing}

Both the human-generated and GPT-generated data need to be processed to an acceptable state. Since the origin of the data differs, the processing techniques applied also differ. 

\subsubsection{Reddit Data Processing}

In order to reduce unwanted bias towards human-generated content, we remove any features that exist in the Reddit data that does not exist in GPT data. Specifically, we remove links and any other non-text multi-modal information from  the Reddit data. Links can exist in both markdown formatting (text)[link] as well as general URL formatting, so we must handle both uniquely. We extract the text from links in markdown formatting and remove the link and special characters. For generic URLs, we simply remove them since GPT3.5 does not generate links.

Special characters that are not representative of typical punctuation are also removed. An example is bold characters in markdown, which are encapsulated by * characters. We also remove newline characters from human-generated content, as they are not present in GPT-generated text. Other bias exists in human-generated content, such as personal anecdotes and nuanced contextual understanding, but said bias can be harnessed to discern human-generated content from GPT-generated content since GPT can attempt to replicate said biases through more advanced prompting techniques.

We filter out Reddit data with profanity or other inappropriate content using the better-profanity library, since GPT does not typically use any profanity or generate inappropriate content unless specifically prompted to do so. The better-profanity library can filter most inappropriate content automatically, but any left over data has been manually filtered. We also filter out any human-generated content under 100 characters in length, as these comments are short and typically lacking in substance.

\subsubsection{GPT Data Processing}

The GPT data requires some minimal processing before it can be utilized. The output of GPT is limited to 100 tokens to match the typical length of Reddit comments so as to avoid any bias in length. Since the token limit can result in incomplete sentences, any incomplete sentences in the GPT responses are removed.
In order to ensure the preservation of the underlying patterns, while also avoiding the introduction of human bias into the GPT data, so we don't perform any further processing.

\section{Proposed Method}

\begin{figure}[!htp]
\begin{center}
\includegraphics[width=0.8\columnwidth]{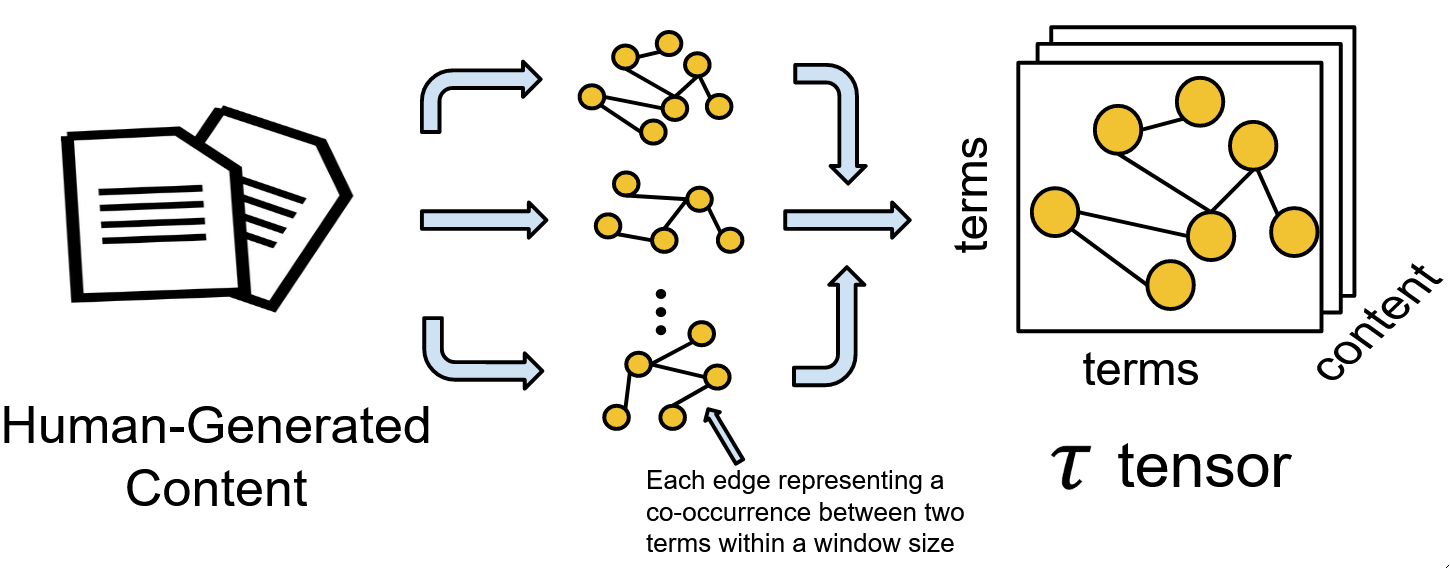}
\caption{Tensor construction method. We generate a unique graph for each document, where each edge represents the co-occurrence between two unique terms within a set window size, and then stack the graphs to create the tensor.} \label{tensor_creation}
\end{center}
\end{figure}

We propose \method, a novel method for anomaly detection that leverages tensor decomposition  to identify underlying patterns in the data. Specifically, we are leveraging the fact that comparing the reconstruction of a tensor built from its decomposed components to the original may show that the original tensor contains anomalies, if a significant difference between the two exists. This type of tensor representation has been successfully applied to two very different language tasks: fake news detection \cite{guacho2018misinformation} and humor recognition \cite{humorrecognition}. So its reasonable that this method can also be applied directly to GPT detection, and we have applied it to the task using this dataset. The proposed method is structured as a pipeline and consists of a few major components, which this section will describe in detail. This approach is considered semi-supervised based on the first step.

The first step of the pipeline is to construct a three-dimensional tensor of the corresponding input data. In most cases, the tensor should represent the in-distribution data. If this is the case, then any positive data points will be excluded from the tensor, hence the semi-supervised approach.

We build the tensor as follows. For each document in the data, we build a co-occurrence matrix for each term and its neighbors within a window size (typically 5 to 10). Each co-occurrence matrix is an $M \times M$ matrix, where M represents the number of unique terms in the entire collection of documents. So, given a collection $C$ with $N$ documents, we will construct an $N\times M \times M$ tensor where each slice of the tensor is an $M \times M$ co-occurrence of document $n_i$; there are $N$ such co-occurrence matrices (Fig. \ref{tensor_creation}). 

An important distinction is that we are only using human-generated content to build the tensor. Since we require labeled non-GPT data in order to build the tensor, we consider this method semi-supervised. Simultaneously, we only use terms from the human-generated content to build the the co-occurrence matrices in order to avoid any potential contamination from the test set.

The second step of the pipeline is to decompose the tensor. For our proposed method, we employ the Canonical Polyadic Decomposition (CPD)\cite{guacho2018misinformation} to decompose the tensor into factor matrices.

\begin{figure*}[!htp]
\includegraphics[width=0.7\textwidth]{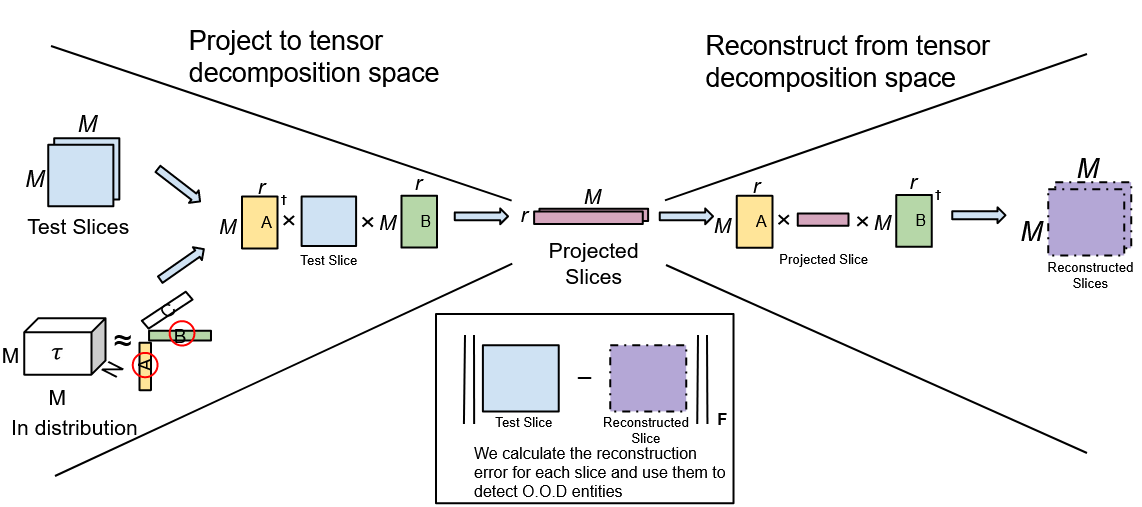}
\caption{Each slice from the test set is projected via the decomposition factors A and B, and then reconstructed. The reconstruction error for each slice is the Frobenius norm between the slice and its associated reconstruction. } \label{reconstruction_method}
\end{figure*}

The third step of the pipeline is to project and reconstruct each slice of the tensor using the decomposed factor matrices. Specifically, we want to construct a vector $r$ of length $N$ which contains the reconstruction error of each slice in the tensor (Fig. \ref{reconstruction_method}). Since the input tensor is three dimensional, CPD decomposition calculates three corresponding factor matrices $\textbf{A}, \textbf{B}$, and $\textbf{C}$ of dimensions $M \times r$, $M \times r$, $N \times r$ respectively, where $r$ represents the rank of the decomposition. We will then project each slice of the tensor, denoted as $\textbf{S}_i$, through the factor matrices $\textbf{A}$ and $\textbf{B}$, to get the projection $\textbf{P}_i$ as  $\mathbf{P}_i = (\textbf{A}^\dagger \cdot \mathbf{S}_i) \cdot \textbf{B}$.
We can then calculate the reconstruction  $\textbf{S}_i'$ for each slice $\textbf{S}_i$ as 
    $\textbf{S}_i' = \textbf{A} \cdot (\mathbf{P}_i) \cdot \textbf{B}^\dagger$
The reconstruction $\textbf{S}_i'$ will be of dimension $M \times M$, which is the same as $\textbf{S}_i$. We can then take the Frobenius norm to calculate the reconstruction error $e_i$ of each slice $\textbf{S}_i$:
$ e_i = || \textbf{S}_i' - \textbf{S}_i ||_F$.

The reconstruction errors follow a distribution which can be modelled by both supervised and unsupervised models, however, given that our method only has access to negative labels (i.e., human-generated text), we employ unsupervised anomaly detection to model the reconstruction error distribution and identify positive (i.e., GPT-generated text) as O.O.D points.

\section{Experimental Evaluation}

\subsection{Baseline methods}
We use three distinct models—Random Forest, Support Vector Machine (SVM), and BERT on the dataset to assess their efficacy in GPT-generated text detection. The selection of these models is intended to explore a spectrum of approaches \cite{crothers2023machine}. 
For both SVM and Random Forest, we applied a simple TF-IDF vectorizer to transform the text data into numerical data\cite{jones1972a}, which is then fed into both models. All results are obtained via 10-fold cross-validation.

BERT-based models stand as state-of-the-art models in natural language processing and represents the cutting edge of deep learning-based text understanding \cite{crothers2023machine}. Its ability to contextualize words within a sentence and grasp intricate semantic relationships makes it an ideal candidate for the task of distinguishing between human-generated and GPT-generated text. For this reason, we have chosen a pre-trained baseline BERT model for experimentation.

\subsection{Results}

Our results indicate that BERT outperformed both SVM and Random Forest in terms of ROC-AUC values (Table \ref{tab:tab1}). This observation aligns with the expectations set by the choice of models, with BERT leveraging its advanced contextual understanding to discern the nuances of GPT-generated text. 

SVM, representing a traditional ML approach, demonstrated respectable performance, while Random Forest, as an ensemble method, showcased its ability to capture certain patterns but fell short of the performance achieved by BERT. Despite being outperformed, SVM and Random Forest remain viable options for GPT-generated text detection due to their interpretable structure and efficiency in handling high-dimensional feature spaces. These traditional machine learning approaches offer transparency in model predictions and can serve as practical alternatives, particularly when computational resources are constrained or interpretability is an important consideration.

\begin{table}[!ht]
\centering
\small
    \caption{Performance metrics on the GPT Detection Dataset. Note that \method{}, while semi-supervised, performs comparably to fully-supervised baselines.}
    \label{tab:tab1}
    \vspace{-0.05in}
    {\begin{tabular}{c|cccc|c}
        \toprule
        \multicolumn{1}{c|}{} & \textbf{F1} & \textbf{AUC}  \\
        \midrule
        \multirow{1}{*}{BERT}     & 0.934 & 0.984 \\
        \midrule
        \multirow{1}{*}{SVM}      & 0.813 & 0.845 \\
        \midrule
        \multirow{1}{*}{Random Forest}    & 0.787 & 0.825 \\
        \midrule
        \multirow{1}{*}{\method{} (proposed)}    & 0.667 & 0.708 \\
        \bottomrule
    \end{tabular}}
\end{table}

For \method{} we compare the results of applying an unsupervised model for anomaly detection on the calculated reconstruction errors (Fig. \ref{reconstruction_method}). We apply models from the PyOD library, a comprehensive Python library for outlier detection \cite{zhao2019pyod}. PyOD supports both supervised and unsupervised models, and in both cases provides a simple abstraction to gather predictions and metrics from the model. For unsupervised models, the anomaly scores given to each data point by the model are converted to a prediction by applying a threshold. Thus, we can compare metrics such as F1-score and ROC AUC score against other supervised models. The best performing unsupervised model from our experiments was the KDE model, which assesses the likelihood of each data point by estimating its probability density function based on a non-parametric approach, identifying anomalies as instances with lower likelihoods.

\section{Conclusions}
In this paper we introduced \dataset, a novel benchmark dataset for the detection of GPT-generated text and demonstrated the performance of fully-supervised methods on it. Furthermore, we proposed \method{}, a tensor-based method which  only has access to human-generated data and is able to perform on par with fully-supervised baselines.

\section{Acknowledgement}
\small{This research was supported by the National Science Foundation under CAREER grant no. IIS 2046086 and CREST Center for Multidisciplinary Research Excellence in Cyber-Physical Infrastructure Systems (MECIS) grant no. 2112650, and by the Combat Capabilities Development Command Army Research Laboratory and was accomplished under Cooperative Agreement Number W911NF-13-2-0045 (ARL Cyber Security CRA). The views and conclusions contained in this document are those of the authors and should not be interpreted as representing the official policies, either expressed or implied, of the Combat Capabilities Development Command Army Research Laboratory or the U.S. Government. The U.S. Government is authorized to reproduce and distribute reprints for Government purposes not withstanding any copyright notation here on.}%

\bibliographystyle{ACM-Reference-Format}
\bibliography{bibliography}

\end{document}